\documentclass{article}
\usepackage{spconf,amsmath,epsfig}
\usepackage[colorlinks,linkcolor=blue]{hyperref}

\usepackage{subfigure}

\usepackage{multirow}
\usepackage{booktabs}
\usepackage{algorithm}
\usepackage{algorithmic}
\usepackage{epstopdf}
\usepackage{color}

\usepackage{amsmath}
\usepackage{amssymb}

\usepackage{makecell}
\usepackage{verbatimbox}

\usepackage[numbers,sort&compress]{natbib}
\begin{document}\sloppy

% Example definitions.
% --------------------
\def\x{{\mathbf x}}
\def\L{{\cal L}}

% Title.
% ------
%\title{DistGAN:Distillation Based GAN for Controllable Face Aging}
\title{Controllable Face Aging}
%
% Single address.
% ---------------
\name{Haien Zeng,  Hanjiang Lai,  Jian Yin}
\address{}
%Address and e-mail should NOT be added in the submission paper. They should be present only in the camera ready paper.
\maketitle

\begin{abstract}
Motivated by the following two observations: 1) people are aging differently under different conditions for changeable facial attributes, e.g., skin color may become darker when working outside, and 2) it needs to keep some unchanged facial attributes during the aging process, e.g., race and gender, we propose a controllable face aging method via attribute disentanglement generative adversarial network. To offer fine control over the synthesized face images, first, an individual embedding of the face is directly learned from an image that contains the desired facial attribute. Second, since the image may contain other unwanted attributes, an attribute disentanglement network is used to separate the individual embedding and learn the common embedding that contains information about the face attribute (e.g., race). With the common embedding, we can manipulate the generated face image with the desired attribute in an explicit manner. Experimental results on two common benchmarks demonstrate that our proposed generator achieves comparable performance on the aging effect with state-of-the-art baselines while gaining more flexibility for attribute control. Code is available at supplementary material. 
\end{abstract}
\section{Introduction}
\label{sec:intro}

Face aging~\cite{fu2010age}, also called face age progression, is to synthesize faces of a person under different ages. It is one of the key techniques for a variety of applications, including looking for the missing children, cross-age face analysis, movie entertainment and so on. 

Considerable research effort~\cite{yang2018learning,shu2015personalized,wang2016recurrent,he2019s2gan,liu2019attribute,zhang2017age} has been devoted to generating realistic aged faces, in which deep generative adversarial network (GAN) has become one of the leading approaches in face age progression. For example, Zhang \textit{et al.}~\cite{zhang2017age} proposed conditional adversarial autoencoder framework (CAAE) for age progression,
% and Liu et al.~\cite{} present a contextual generative adversarial nets for face aging. Preserving the personalized factors is very important to synthesize the authentic and rational face images~\cite{he2019s2gan} during the aging process. Hence, 
and Wang \textit{et al.}~\cite{wang2018face} proposed identity-preserved conditional generative adversarial network, in which the perceptual loss is introduced to keep identity information. Facial attributes are very important to synthesize the authentic and rational face images~\cite{lu2018attribute}. For the changeable attributes, they may change in different conditions. For example, working conditions, chronic medical conditions and lifestyle habits may affect individual aging processes. For the unchanged facial attributes, some researchers
have found that the unpaired training data may lead to unnatural changes of facial attributes ~\cite{liu2019attribute}.
%e.g., Liu \textit{et al.}~\cite{liu2019attribute} proposed an attribute-aware face aging model via the wavelet-based GAN. 
Motivated by that, we propose a controllable face aging to synthesize face images with the desired attributes (e.g., race and gender). 

In parallel, style-based generative adversarial network~\cite{karras2019style,gatys2016image}, which render the content image in the style of another image, has drawn much interest. The style-based generator can not only generate impressively high-quality images but also offer control over the style of the synthesized images~\cite{liu2019few,huang2017arbitrary}.
% These style-based GAN can render the content image in the style of another image~\cite{gatys2016image}. Motivated by that, we design a style-based generator architecture for controllable face aging. For example, we can synthesize elderly face image with the desired attributes (e.g., white male) by proving a style image of a white male. 
However, the style-based GAN has two limitations. First, it transfers the style from one image to another, but not the common age pattern~\cite{wang2018face}. Second, the style images may contain other unwanted facial attributes, such as eyeglasses and smiling. If we use these style images, the style-based GAN may generate the face images contain other unwanted attributes (please refer to SubSection~\ref{section_ablation_study} for more details). Hence, it is still hard to control the style of the synthesized face image if we do not have the perfect style images that only have the desired attributes as inputs.  

In this paper, we propose an attribute disentanglement GAN for controllable face aging. Our framework is based on the style GAN as shown in Figure~\ref{main_model}. We first encode an image with the desired attribute into a latent and individual embedding. Instead of directly using that individual embedding, we learn an attribute disentanglement network, which disentangles and distils the knowledge of the individual embedding. This process aims to learn the common embedding that contains information about the age pattern of all images with facial attribute. %This process can depress the unwanted attributes with a less number occurring and obtain more reliable attribute embedding.
Finally, we feed the learned common embedding to the decoder via the adaptive instance normalization (AdaIn)~\cite{huang2017arbitrary} blocks, which can control the aging process with the desired attribute in an explicit manner.

The main contributions of our work are three-fold. First, we propose an attribute disentanglement GAN for controllable face aging. Compared to the existing GAN-based face aging models, our method offers more control over the attributes of the generated face images at different levels. Second, we propose an attribute disentanglement method to obtain more reliable age pattern, which can remove the effect of other unwanted attributes. Finally, we qualitatively and quantitatively evaluate the usefulness of the proposed method. 

\section{Related Work}
\textbf{Face Aging} \@ \@ \@ Traditional face aging approaches~\cite{fu2010age,shu2015personalized} can be divided into two categories, physical model-based approaches and prototype-based methods. The physical model-based methods~\cite{suo2012concatenational} focus on modelling the aging mechanisms, e.g., skin's anatomy structure and muscle change. The prototype-based methods~\cite{kemelmacher2014illumination} use the differences between the average faces of age groups as the aging pattern.
Recently, generative adversarial network~\cite{goodfellow2014generative} based methods have also been widely studied for solving the aging progression problems. For example, Wang \textit{et al.}~\cite{wang2016recurrent} introduced a recurrent neural network for face aging.  Antipov \textit{et al.} \cite{antipov2017face} proposed to apply conditional GAN for face aging, and \cite{zhang2017age} is an auto-encoder conditional GAN. Song \textit{et al.}~\cite{song2018dual} present a dual conditional GANs for simultaneously rendering a series of age-changed images of a person. In~\cite{yang2018learning}, it is a pyramid architecture of GANs for aging accuracy and identity permanence. Although tremendous progress have been made by GAN-based methods, limited attention has paid for face aging with controllable attributes. In this paper, we propose a user-controllable approach for face aging. 

\textbf{Style Transfer} \@ \@ \@  Another similar work is style transfer~\cite{karras2019style,johnson2016perceptual,gatys2016image}, which synthesizes image whose contents are from one input image and the style is from another artistic style image. Gatys \textit{et al.}~\cite{gatys2016image} firstly used the feed-forward neural network to obtain impressive style transfer results. Lately, Johnson \textit{et al.}~\cite{johnson2016perceptual} proposed to use the perceptual loss functions for training a feed-forward network. A novel adaptive instance normalization (AdaIn) layer~\cite{huang2017arbitrary}, which simply adjusts the mean and variance of the content input to match the style of another input, is able to do arbitrary style transfer in real-time. Inspired by that, Karras et al.~\cite{karras2019style} propose a style-based GAN to control the image synthesis process. However, the style transfer focus on transferring the style of one image to another image, while face aging requires to transfer the age pattern to input face~\cite{wang2018face}. Hence, it is hard to directly applied the style transfer to face aging. In this paper, we propose a style disentanglement module, which can remove the unwanted attribute and learn the common age pattern, making the available style transfer for face aging.

\section{Proposed method}
\label{method}

\begin{figure*}[ht]
  \centering
    \includegraphics[width=0.9\hsize \hspace{0.01\hsize}]{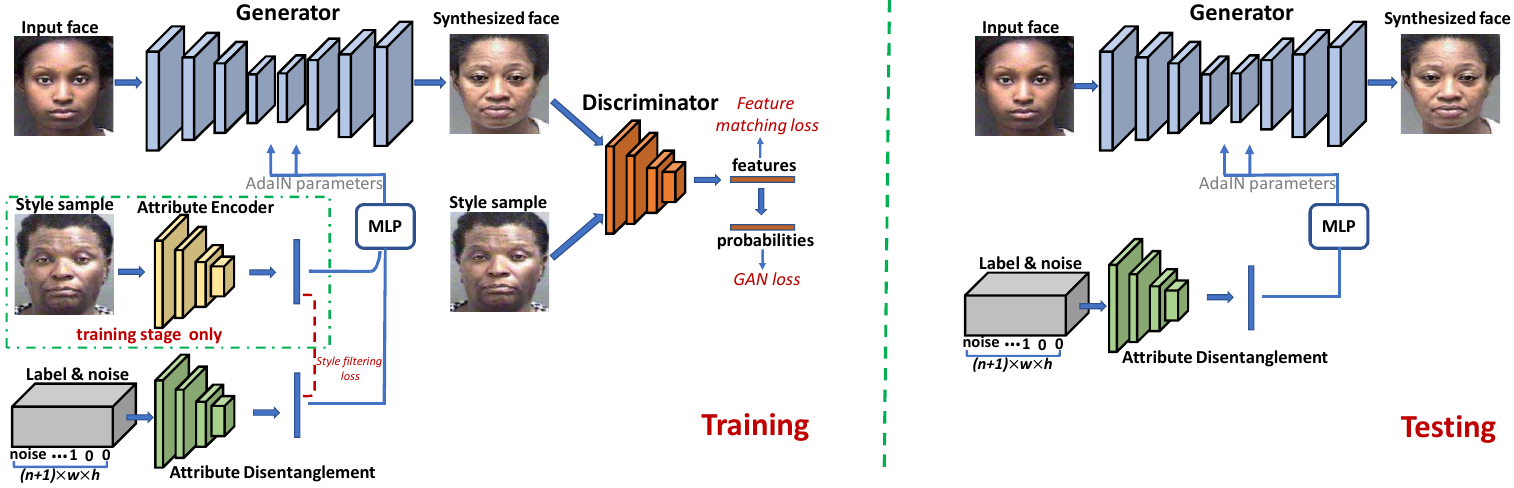}
  \caption{Model structure of our method. The left is the structure in training stage while the right is for testing.}
  \label{main_model}
\end{figure*}

\subsection{Overview}
In this paper, we propose an attribute disentanglement GAN for controllable face aging. Similar to the style transfer, our method has two inputs: an input face image and desired attributes, i.e., ages with desired facial attributes. For ease of presentation, we only consider two  most common attributes: race and gender.
%~\footnote{Other attributes are less common in the widely used aging datasets, e.g., MORPH. }. 
Please note that our method can be easily extended to other attributes for more control of aging process. Let $S=\{ age, gender, race\}$ to represent the label of the desired attributes, e.g., $S_i = \{21 \sim 30, male, white\}$ represents a white male in $21 \sim 30$ years old. We denote the  face dataset as $[I_i =(X_i,S_i)]_{i=1}^{M_I}$, where $X_i$ is the $i$-th input image and $S_i$ is the corresponding label, $M_i$ is the number of training face images. 

In testing, we use a generator $G$ and an attribute disentanglement $F$ to render the face image $X_i$ with aging effects in the target attributes $S_t$ as shown in the right side of Fig.\ref{main_model}.

In training, we introduce an individual attribute encoder $E$ and a discriminator $D$ for training the $G$ and $F$. Please note that $E$ is used to obtain the individual age pattern while $F$ learns the common age pattern. To train attribute encoder $E$, we also introduce the style images as $[T_t =(X_t,S_t)]_{t=1}^{M_T}$, which contain the desired attributes. The attribute encoder $E$ maps the style image $X_t$ into the individual embedding as $z_t = E(X_t)$. The discriminator $D$ is used to distinguish the generated face images from the real face images. We describe each in detail in the following. 

\textbf{Generator} \@ \@ \@ Fig.~\ref{main_model} show the architecture of the generator. Similar to other GAN-based methods for face aging, it firstly uses multiple down-sampling convolutional layers to learn the high-level feature maps, and then the feature maps go through multiple up-sampling convolutional layers to generate the output image. The main different is that we learn an affine transformation before the up-sampling layers to encode the attribute embedding (e.g., $z =F(S_t)$ or $z = E(X_t)$) into the generator via the AdaIN~\cite{huang2017arbitrary} operations. More precisely, suppose that $f \in \mathbb{R}^{H_f \times W_f \times C_f}$ represents feature maps before a up-sampling layer, where $H_f, W_f, C_f$ are the height, weight, channel, respectively. $f_c \in \mathbb{R}^{H_f \times W_f}$ denotes as the $c$-th channel. For each channel, the AdaIn operates as
\begin{equation}
 \setlength\abovedisplayskip{4pt}
 \setlength\belowdisplayskip{4pt}
 \text{AdaIn}(f_c, z) = z^{\mu} \ \frac{f_c - \mu(f_c)}{\sigma(f_c)} + z^{b},
\end{equation}
where $\mu(f_c)$ and $\sigma(f_c)$ are the mean and standard deviation, $z = (z^{\mu},z^{b})$ is the attribute embedding, e.g., $z$ is the output of the $F$ or $E$. For ease presentation, we denote as  $\hat{X}_{i,t} = G(X_i, F(S_t))$, where $\hat{X}_{i,t}$ is the synthesized face image, $X_i$ is an input image, $z = F(S_t)$ encodes the attributes $S_t$ into the common embedding of age pattern. 

\textbf{Attribute Encoder} and \textbf{Attribute Disentanglement} \@ \@ \@ The attribute encode $E$ takes an individual style image $X_t$ as input and encode it into an individual embedding $E(X_t)$. Please Note that it only works in the learning stage and can be removed after finishing training procedure. The  $F$ learns the age pattern directly from the attributes. We use $S \in \mathbb{R}^{n \times w \times h}$ to represent the attributes and the input of the $F$, where $w$ and $h$ are the weight and heigh of the input images and $n$ is the number of attributes. For examples, suppose that there are $n_a$ age groups, $n_g$ genders and $n_c$ races, we have $n = n_a \times n_g \times n_c$. We use one-hot code for the attributes, in which only one feature map is filled with one and others are filled with zero. To generate  more diverse images, we also add a noise channel in $S$, and finally obtain a code $S \in \mathbb{R}^{(n+1) \times w \times h}$. The attribute disentanglement takes $S$ as input and obtains common embedding $F(S_t)$ with the help of the $E$.

\textbf{Discriminators} \@ \@ \@ Following~\cite{liu2019few}, we use multiple binary classifications (each binary classification for one attribute) instead of one multi-classification problem. Liu \textit{et al.}~\cite{liu2019few} showed that multiple binary classifications perform better than a hard multi-class classification. We denote $D^{S_t}$ as the $t$-th binary adversarial classification for the attribute $S_t$, which distinguishes the generated face images from the real images.

\subsection{Training}
In this paper, we propose a two-stage method to alternatively train the four modules. In the first stage, we learn $G,E$ and $D$ three modules, which learns to render the input face with the facial attributes of another face image. In the second stage, we further learn the attribute disentanglement $F$, which disentangles the unwanted attributes in the individual face image and obtain the common age pattern. 

\textbf{Updating $G,E,D$ via individual attribute translation.} 
There are two input images: one input face $X_i$ and one style image $X_t$, with their corresponding labels are $S_i$ and $S_t$, respectively. In this stage, we learn to render the input face image $X_i$ in the style of another face image $X_t$, similar to the existing style-based GAN. We follow \cite{liu2019few} and the loss function is denoted as
\begin{equation}
\setlength\abovedisplayskip{2pt}
\setlength\belowdisplayskip{-3pt}
\max_{D} \min_{G,E} \ell_{GAN}(D,E,G) + \lambda_1 \ell_R(G,E) + \lambda_{2}\ell_{FM}(G,E),
\end{equation}
where the GAN loss $\ell_{GAN}(D,E,G)$ is formulated as
\begin{equation}
\setlength\abovedisplayskip{4pt}
\setlength\belowdisplayskip{4pt}
\begin{aligned}
&\ell_{GAN}(D,E,G) = \mathbb{E}_{\{ X_i \} }[\text{log}D^{S_i}(X_i)] & \\
& + \mathbb{E}_{\{ X_i,X_t \} }[\text{log}(1 - D^{S_t}(G(X_i, E(X_t)))],&
\end{aligned}
\end{equation}
and the reconstruction loss $\ell_R(G,E)$ is defined as
\begin{equation}
\setlength\abovedisplayskip{4pt}
\setlength\belowdisplayskip{4pt}
\ell_R(G,E) = \mathbb{E}_{\{ X_i \}} [||X_i - G(X_i, E(X_i))||_1^1],
\end{equation}
where the $\ell_R(G,E)$ loss require to reconstruct the input image when both the input face image and style image are the same image. 

The feature matching loss $\ell_{FM}(G,E)$ learns to minimize the features' distance, in which the two features are extracted from the output image $\hat{X} = G(X_i, E(X_t))$  and the style image $X_t$. We use the second-last layer of discriminator $D$, denoted as $D_f$, to extract the features. It is used to regularize the training~\cite{liu2019few}, which is formulated as 
\begin{equation}
\setlength\abovedisplayskip{4pt}
\setlength\belowdisplayskip{4pt}
\ell_{FM}(G,E) = \mathbb{E}_{\{ X_i,X_t \}} [|| D_f(\hat{X} ) - D_f(X_t)||_1^1].
\end{equation}

\textbf{Updating $F$ via common attribute translation.} We fixed the parameters of $E,G,D$ and update the attribute disentanglement $F$. The optimization object is defined as
\begin{equation}
\setlength\abovedisplayskip{2pt}
\setlength\belowdisplayskip{-5pt}
\begin{aligned}
\min_{F} & \ \ \ \ell_{GAN}(D,F,G) + \lambda_1 \ell_R(G,F) + \lambda_{2}\ell_{FM}(G,F)& \\
& + \lambda_3 \ell_{DIS}(E,F),&
\end{aligned}
\end{equation}
where the first three terms are the same as the first stage except that we use $F(S_t)$ instead of $E(X_t)$ as the attribute embedding. That is in the first stage the output of the generator is $\hat{X} = G(X_i, E(X_t))$, now it becomes   $\hat{X} = G(X_i, F(S_t))$. To transfer the knowledge from the $E$ to $F$, we introduce a new disentanglement loss $ \ell_{DIS}(E,F)$, which is defined as
\begin{equation}
\setlength\abovedisplayskip{4pt}
\setlength\belowdisplayskip{4pt}
 \ell_{DIS}(E,F) = ||\hat{X}_{E} - \hat{X}_F ||_1^1 +
 \beta ||E(X_t) - F(S_t) ||_1^1,
\end{equation}
where $\hat{X}_E = G(X_i, E(X_t))$ and  $\hat{X}_F = G(X_i, F(S_t))$. 

Please note that there are many style images $X_t$ and one attribute conditional $S_t$, thus, $F(S_t)$ can learn the common age pattern of all style face images. This process can depress the unwanted attributes and obtain more reliable common embedding.

\begin{figure*}[th]
  \vspace{-0.45cm}
  \centering
    \includegraphics[width=0.7\hsize \hspace{0.01\hsize}]{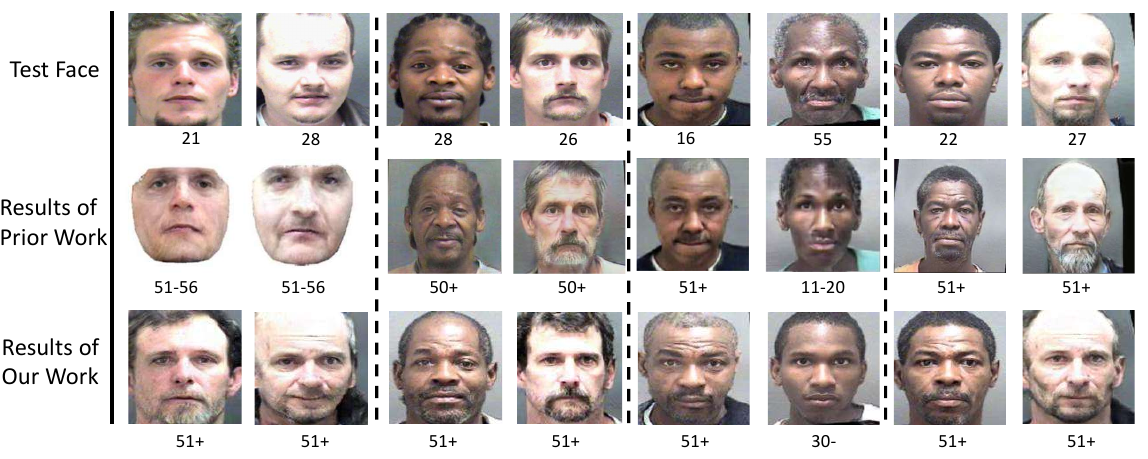}
  \caption{Comparison with prior work (the second row) of age progression. Four methods are considered and two sample results are presented for each. These four methods are: HFA \cite{yang2016face}, PAG-GAN \cite{yang2018learning}, IPCGAN \cite{wang2018face} and Attribute-aware GAN~\cite{liu2019attribute}.}
  \label{comparison}
\end{figure*}

\begin{figure*}[t]
%   \vspace{-0.5cm}
%   \vspace{-1cm}
  \centering
    \includegraphics[width=0.7\hsize \hspace{0.01\hsize}]{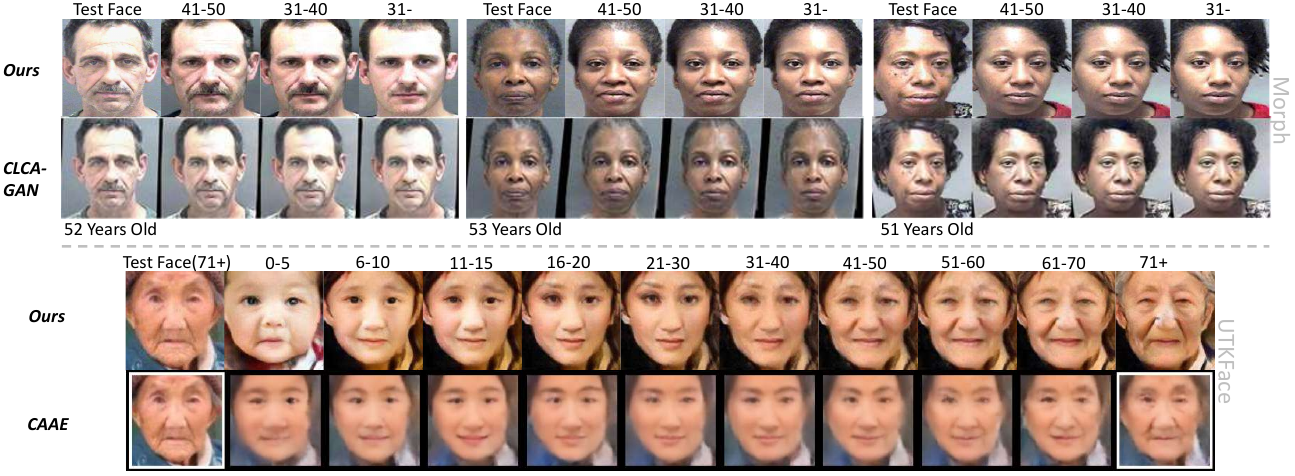}
  \caption{Age regression results on Morph and UTKFace. We compare our results with GLCA-GAN~\cite{li2018global} and CAAE~\cite{zhang2017age}, respectively.}
  \label{regression}
\end{figure*}

\section{Experiments}
\subsection{Datasets and Implementation Details}
\textbf{MORPH}~\cite{ricanek2006morph} is a large-scale face dataset, which consists of 55,134 images of more than 13,000 individuals. There are an average of 4 images per individual, and the ages range from 16 to 77.  Following the settings in \cite{liu2019attribute}, we divide the images of MORPH into four age groups, i.e., 30-, 31-40, 41-50, 51+. 

\textbf{UTKFace}~\cite{zhang2017age}~\footnote{https://susanqq.github.io/UTKFace/} consists of over 20,000 face images with annotations of age, ethnicity and gender. Their ages range from 0 to 116 years old.  We follow the setting in~\cite{zhang2017age} and divide face images into ten age groups, i.e., 0-5, 6-10, 11-15, 16-20, 21-30, 31-40, 41-50, 51-60, 61-70, and the rest. 

In our experiments, all images are resized to be $128 \times 128$ and the RMSProp is chosen to be the optimizer. The learning rate and batch size are setted to be $1e^{-4}$ and $10$, respectively. The maximum iterations in the two stages are set to 100,000 and 50,000, respectively.  The parameters in the proposed architecture are all initialized with the kaiming initialization. Due to the space limitation, more details of the implementation can be found in the supplementary material, in which we provide the \textbf{source code} for our implementation.

\subsection{Comparison with Prior Work}
In this set of experiments, we compare the performance of the proposed method with several state-of-the-art prior work: HFA~\cite{yang2016face}, GLCA-GAN~\cite{li2018global}, CAAE~\cite{zhang2017age}, IPCGAN~\cite{wang2018face}, PAG-GAN~\cite{wang2018face} and Attribute-aware GAN~\cite{liu2019attribute}. Some baselines aim to learn the unchanged facial attributes for preserving the identity information, e.g, the attribute-aware GAN~\cite{liu2019attribute}.
%For example, the attribute-aware GAN~\cite{liu2019attribute} embeds the facial attribute vectors into both the generator and discriminator to preserve the unchanged facial attributes, e.g., gender and race. 
To make a fair comparison, we show that the proposed method can generate high-quality face images and also preserve the unchanged facial attributes. For example, given an input $(X_i, S_i)$ with $S_i = \{age_i, gender_i, race_i\}$ and the target age is $age_t$, then we can synthesize the age face image via $G(X_i, F(S_t))$ where $S_t = \{age_t, gender_i, race_i\}$ as shown in Fig.~\ref{main_model}. 

\begin{table}[ht]
\small
\centering
\caption{Complexity analysis of different methods, where $n_a$ is the number of age groups.}
\label{table1}
% \begin{tabular}{|c|c|c|c|c|c|c|c|c|c|c|c|c|}
%\begin{tabular}{|c|p{0.9cm}p{0.9cm}p{0.9cm}p{0.9cm}|}
\begin{tabular}{|p{3cm}<{\centering}|p{3cm}<{\centering}|}
\hline
%\multirow{2}{*}{{Method}}&\multicolumn{4}{c|}{IAPR TC-12}\\
%\cline{2-5}
Method & \# Models \\
\hline
Wang et al. \cite{wang2018face} &1  \\
Li et al. \cite{li2018global} &1  \\
Liu et al. \cite{liu2019attribute} &$n_a(n_a+1)$  \\
Yang et al. \cite{yang2018learning} &$n_a(n_a+1)$  \\
\hline
Ours &1  \\
\hline
\end{tabular}
\label{complexity}
\end{table}

\begin{figure*}[t]
  \vspace{-0.4cm}
  \centering
    \includegraphics[width=0.7\hsize \hspace{0.01\hsize}]{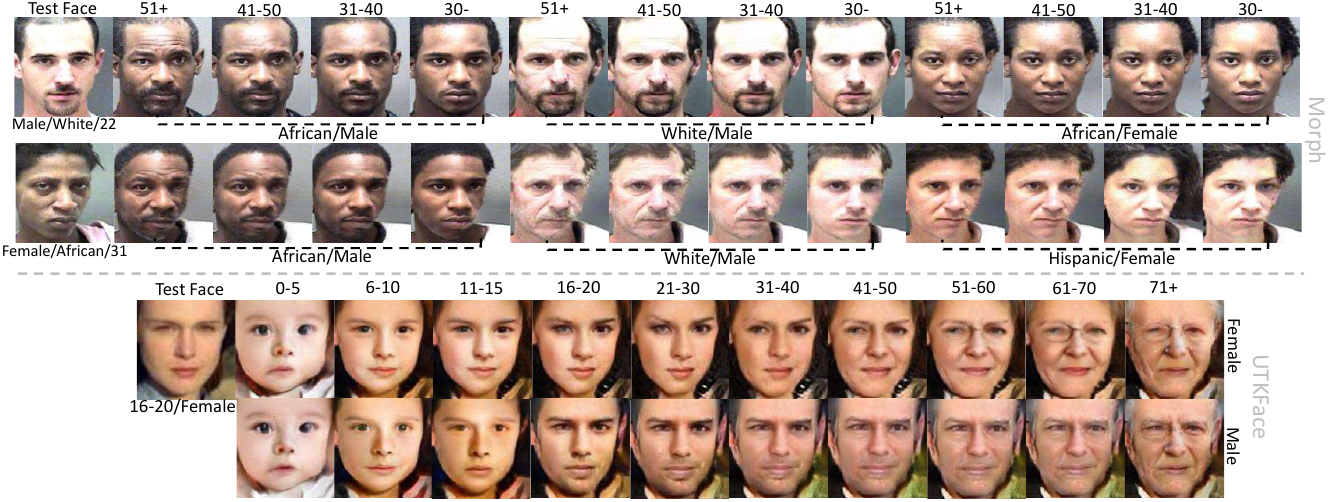}
  \caption{Results of attributes controllable synthesization experiment on Morph and UTKFace. The First column is the test faces while others are synthesis results on the conditions of different races or genders.}
  \label{controllable}
\end{figure*}

\begin{figure*}[t!]
  \centering
    \includegraphics[width=0.81\hsize \hspace{0.01\hsize}]{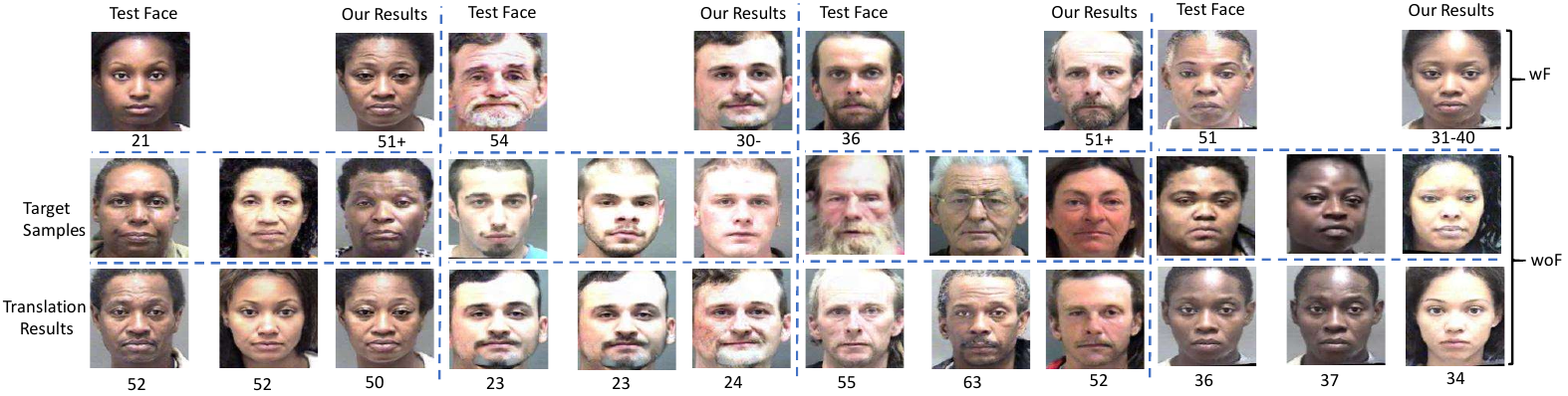}
  \caption{Ablation study results of 4 different individuals. The first row is the test faces and output results of the model with style disentanglement $F$. The second row is the style images that required by the model without style disentanglement $F$ and the last row is the output results of the model without $F$.}
  \label{ablation_study}
\end{figure*}

The comparison results are shown in Fig.~\ref{comparison} and  Fig.~\ref{regression}. We can see that our method outperforms some baselines, e.g, HFA, CAAE and GLCA-GAN, and achieves comparable results with the PAG-GAN and Attribute-aware GAN. These comparison results show that the proposed method can also generate the high-quality face images. Table~\ref{complexity} shows the complexity analysis of different methods. Our method only needs one model to preserve the facial attributes for $n_a$ age groups, while the PAG-GAN and Attribute-aware GAN need to train $n_a(n_a+1)$ models. Form the above results, we can see that our method achieves comparable results while gaining more flexibility for attribute control. 

\textbf{Facial Attribute Consistency}  \@ \@ \@ We also evaluate the performance of facial attributes preservation which follows the settings of~\cite{liu2019attribute}. We randomly sample 2,000 images to compute the preservation rate. The results of competitors are directly cited from~\cite{liu2019attribute} for fair comparison.  The comparison results are shown in Table~\ref{gender} and Table~\ref{race}. We can see that our method performances better than the baselines for preserving the facial attributes.

\begin{figure*}[t]
  \centering
    \includegraphics[width=0.73\hsize \hspace{0.01\hsize}]{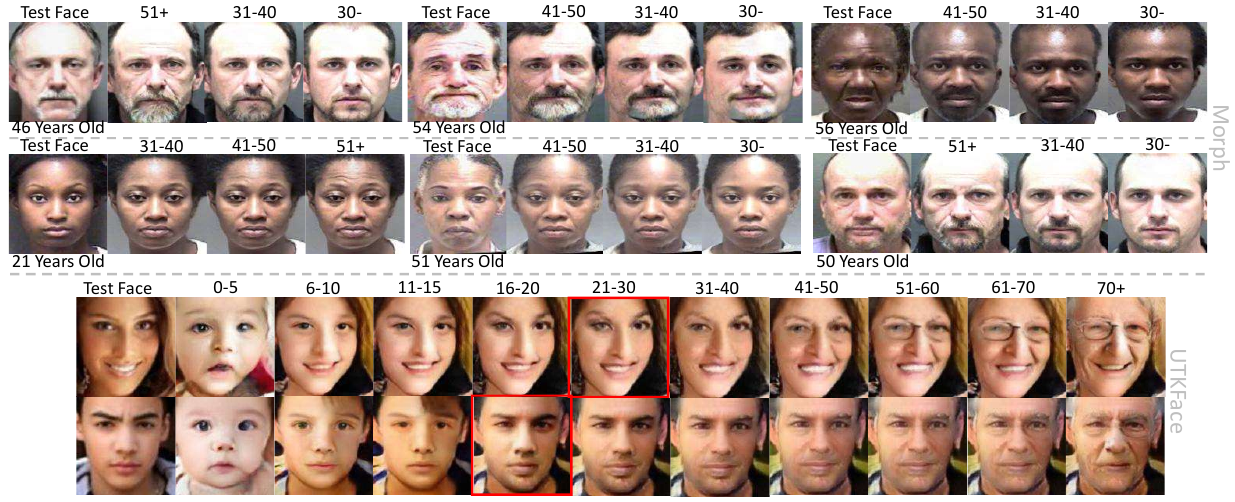}
  \caption{Sample results on Morph and UTKFace. The first two rows are results of Morph while the last two rows are for UTKFace. The red boxes in the results of UTKFace indicate the age groups of the input face images.}
  \label{samples}
\end{figure*}

We also show some results in Fig.~\ref{samples}. Besides, to demonstrate the controllability and flexibility of our method, we generate the face images with different attributes as shown in Fig.\ref{controllable}. In our method, face images of different attributes can be synthesized via only changing the condition attribute $S_t$, e.g., European to African, male to female, young to old, and so on. As can been seen that our model is able to conditionally synthesize different face images of different races, ages and genders. 
 
\begin{table}[ht]
\small
\centering
\caption{Facial attributes preservation rate for `Gender'.}
\label{table1}
\begin{tabular}{|c|c|c|c|}
\hline
Method & 31-40 & 41-50 & 51+ \\
\hline
Yang et al. \cite{yang2018learning} &95.96 &93.77 &92.47 \\
Liu et al. \cite{liu2019attribute} &97.37 &97.21 &\textbf{96.07} \\
Ours &\textbf{97.50} &\textbf{97.43} &95.25  \\
\hline
\end{tabular}
\label{gender}
\end{table}

\begin{table}[ht]
\small
\vspace{-0.5cm}
\centering
\caption{Facial attributes preservation rate for `Race'.}
\label{table1}
\begin{tabular}{|c|c|c|c|}
\hline
Method & 31-40 & 41-50 & 51+ \\
\hline
Yang et al. \cite{yang2018learning} &95.83 &88.51 &87.98 \\
Liu et al. \cite{liu2019attribute} &95.86 &94.10 &93.22 \\
Ours &\textbf{96.55} &\textbf{95.75} &\textbf{95.60}  \\
\hline
\end{tabular}
\label{race}
\end{table}

\subsection{Ablation Study}\label{section_ablation_study}
In this set of our experiments, we do ablation study to clarify the impact of the proposed attribute disentanglement $F$ on the final performance. Without the attribute disentanglement, it requires two inputs: an input image $X_i$ and a style image $X_t$. Then, we can generate the image as $G(X_i, E(X_t))$. 

Fig.~\ref{ablation_study} shows some examples, where the second row is the style images and the third row is the generated images $G(X_i, E(X_t))$ that take the test face and the style image as inputs, respectively. Two observations can be observed. 1) The generated face images of $G(X_i, E(X_t))$ always look like their style images, e.g., the second column and the last column. While our proposed attribute disentanglement $G(X_i, F(S_t))$ can learn the common age pattern, e.g, the first row, which can make the style transfer available for face aging. 2) The generated face images without $F$ may contain other unwanted facial attributes. For example, in the penultimate column, even the input image and the style image belong to the same race and gender, the skin color of the generated face is not the wanted attributes. And our proposed method can solve the problem and provide fine control over the generated face images.

\section{Conclusion}
In this paper, we proposed a controllable face aging method based on the  attribute disentanglement generative adversarial network. In the proposed aging architecture, a face image and desired attributes are used as inputs. Then we proposed attribute encoder and  attribute disentanglement two modules to learn the latent embedding that contains the common age pattern of the desired facial attributes. Finally, we used the adaptive instance normalization layer to render the input image with the style of the common embedding.  The experimental results showed that our proposed method can achieve comparable performance with more flexibility for attribute control. 

\bibliographystyle{IEEEbib}
\bibliography{icme2020template}

\end{document}